\documentclass[runningheads]{llncs}

\usepackage[T1]{fontenc}
\usepackage{graphicx}
\usepackage[misc]{ifsym}
\usepackage{hyperref}
\usepackage{multirow}
\usepackage{multirow}
\usepackage{xcolor}
\usepackage{siunitx,etoolbox}

\usepackage{amsfonts}       % blackboard math symbols
\usepackage{nicefrac}       % compact symbols for 1/2, etc.
\usepackage{microtype}      % microtypography
\usepackage[linesnumbered,ruled,vlined]{algorithm2e} 
\usepackage{macros}
\usepackage{color}

\begin{document}
\title{Optimization of Annealed Importance Sampling Hyperparameters}
\author{Shirin Goshtasbpour\inst{1,2}\orcidID{0000-0002-4902-165X} \\\and
Fernando Perez-Cruz\inst{1, 2}\orcidID{0000-0001-8996-5076}}
\authorrunning{S. Goshtasbpour et al.}
\institute{Computer Science dep., ETH Zurich, Ramistrasse 101, 8092 Zurich, Switzerland \and
Swiss Data Science Center, Turnerstrasse 1, 8092 Zurich, Switzerland\\
\email{shirin.goshtasbpour@inf.ethz.ch}}

\toctitle{Optimization of Annealed Importance Sampling Hyperparameters}
\tocauthor{Shirin~Goshtasbpour}
\maketitle              % typeset the header of the contribution
\begin{abstract}
Annealed Importance Sampling (AIS) is a popular algorithm used to estimates the intractable marginal likelihood of deep generative models. Although AIS is guaranteed to provide unbiased estimate for any set of hyperparameters, the common implementations rely on simple heuristics such as the geometric average bridging distributions between initial and the target distribution which affect the estimation performance when the computation budget is limited. In order to reduce the number of sampling iterations, we present a parameteric AIS process with flexible intermediary distributions defined by a residual density with respect to the geometric mean path. Our method allows parameter sharing between annealing distributions, the use of fix linear schedule for discretization and amortization of hyperparameter selection in latent variable models. %Optimization of the bridging distributions with Metropolis-Hasting (MH) correction steps is performed with partial reparameterization and control variates. 
We assess the performance of Optimized-Path AIS for marginal likelihood estimation of deep generative models and compare it to compare it to more computationally intensive AIS.

\end{abstract}
\section{Introduction}
Deep generative models, developed over a decade, are now capable of simulating complex distributions in high dimensional space and synthesizing high quality samples in various domains such as natural images, text and medical data. Many of these models are built with the assumption that data $x\in\X$ resides close to a low dimension manifold. In this case, data can be represented using latent variable $z$ drawn from prior distribution on $\R^d$ with density $p(z)$ and the observation model $p(x|z) = p(x|f_\theta(z))$ is given by a parametric mapping $f_\theta:\R^d\to\Gamma$ where $\Gamma$ is the space of parameters of the likelihood $p(x|z)$.

For complex $f_\theta$ mappings, evaluation of log marginal likelihood 
\begin{align}
\log p(x) = \log \int p(z)p(x|z)dz\nonumber
\end{align}
is intractable and the posterior density $p(z|x) \propto p(x|z)p(z)$ is known only up to a constant normalization factor\footnote{We use normalization factor and marginal likelihood $p(x)$ interchangeably in this paper}. In this case, latent variable inference, probabilistic model evaluation and model comparison are performed using variational approximation of the posterior \cite{blei2017variational,kingma2013auto,rezende2014stochastic} or via sampling methods like Markov Chain Monte Carlo (MCMC) \cite{brooks2011handbook}, Nested sampling \cite{skilling2006nested,naesseth2015nested}, Sequential Monte Carlo (SMC) \cite{del2006sequential} and Annealed Importance Sampling (AIS) \cite{neal2001annealed}. In Variational Inference (VI) the posterior is approximated with the most similar probability density from the family of distributions $\Q = \{ q_\phi(z|x): \phi \in \mathbb{\phi}\}$ indexed by parameter $\phi$. Success of VI depends on sufficient expressivity of members of $\Q$ and our ability to find the optimal member of this family by, for instance, maximizing the Evidence Lower BOund (ELBO)
\begin{align}\label{eq:ELBO}
    \ELBO(\phi) = \int \log \left(\frac{p(x,z)}{q_\phi(z|x)}\right) q_\phi(z|x)dz \leq \log p(x)
\end{align}
where the equality happens only when $q_\phi(z|x) = p(z|x) \in \Q$.

In MCMC we use Markov kernels with unique stationary distribution $p(z|x)$ and sample a Markov chain $\left(z_k\right)_{k\in[M]}$ by iterative application of this kernel on an initial particle $z_0\sim q_\phi(z|x)$. Contrary to parametric VI, under mild assumptions, it is theoretically guaranteed that distribution of $z_M$ converges to the target distribution $p(z|x)$ as $M$ goes to infinity. However, if the posterior has multiple modes or heavy tails, convergence can require large number of iterations and therefore, be computationally prohibitive. Among the sampling methods, SMC and AIS are of particular interest as they produce unbiased estimation of marginal likelihood regardless of the computational budget by assigning importance weights to samples \cite{salakhutdinov2008quantitative,wu2016quantitative,naesseth2018variational}. In these algorithms an auxiliary sequence of distributions, $\left(\pi_k\right)_{k\in[M]}$, is used to bridge a simple proposal distribution with density $\pi_0(z) = q_\phi(z|x)$ and the target density $\pi_M(z) = p(z|x)$. This sequence is defined via unnormalized densities $\left(\gamma_k\right)_{k\in [M]}$ where $\pi_k(z) = \gamma_k(z)/Z_k$ and $\gamma_0(z) = \pi_0(z)$ and $\gamma_M(z) = p(x, z)$. The algorithm produces $N$ Markov chain samples $\left(z_k^j\right)_{k\in[M]}$ for $j\in[N]$ and their corresponding importance weights denoted by $w^j$ as follows: Initially $z_0^j$ is sampled from $\gamma_0$. Then $z_k^j$ is approximately sampled from $\gamma_k$ using a Markov kernel (typically a gaussian distribution around $z_{k-1}^j$ or a transition with invariant distribution $\gamma_k$)\footnote{SMC has an additional resampling step to draw exact samples from $\gamma_k$ to reduce the variance of importance weight although it sometimes results in insufficient sample diversity}. The marginal likelihood is approximated with Monte Carlo method
\begin{align}\label{eq:montecarlo}
    p(x) = \E[w] \approx \hat p^N(x) \coloneqq \frac{1}{N}\sum_{j\in[N]}w^j
\end{align}
where the expectation is taken over joint distribution of the Markov chains. Variance of $\hat p^N(x)$ depends on the selected density sequence and hyperparameters especially when computation resources are scarce. However, little work is available on optimization of the intermediary distributions. In this paper, we study the impact of optimization of AIS hyperparameters for more accurate estimation of log marginal likelihood with only a few annealing distributions. We optimize the sequence of distributions between proposal and target distributions as well as the hyperparameters of Markov kernels with commonly used tuning and training measures to improve the sampling performance in low budget scenarios. We have made the following contributions:
\begin{itemize}
\item We propose the parameterization of a continuous time density path between the proposal and target distributions which can define general density functions. Contrary to \cite{kingma2013auto,kingma2016improved} the densities used do not need to be normalized and we don't require their exact samples since we use sampling algorithms to gradually transition between the intermediary distributions.
\item To optimize the bridging distributions we minimize Jefferys and inverse KL divergences between the distributions of AIS process and its auxiliary reverse process (defined in Section~\ref{sec:AIS}). We empirically show that Jefferys divergence captures both bias and variance of the estimation in benchmark data distributions while inverse KL divergence is unable to do so. 
\item We further implement evaluate our method on deep generative models with different training procedures and achieve comparable results to more computationally expensive version of AIS algorithm.
\end{itemize}

%The combination of both procedures allows reducing the variance of the importance sampling weights providing more accurate estimates with fewer particles and intermediary distributions.
%%%% We illustrate that with optimized hyperparameters we are able to sample from 2d benchmark distributions with fewer number of bridging distributions and compare sample reconstruction and log-likelihood estimation for Variational AutoEncoder \cite{VAE} and Generative Adversarial Networks \cite{GAN}.
The rest of this paper is organized as follows: Section~\ref{sec:rev} is dedicated to reviewing vanilla AIS algorithm. % and BiDirectional Monte Carlo (BDMC) bound.
In Section~\ref{sec:motiv} we present our parameterization of AIS process which results in flexible bridging distributions and motivate out optimization objective. In Section~\ref{sec:optim} we restate a reparameterization method and derivation of the objective gradient estimates previouly developed in \cite{thin2020metflow,thin2021monte}. Finally, we analyze the accuracy of marginal likelihood estimation and its variance on synthetic and image data% and use our method to compare state-of-the-art generative models with implicit likelihood 
in Section~\ref{sec:exp}.

\section{Background}\label{sec:rev}
In this Section we give a brief introduction to AIS algorithm and its popular adaptive versions. %Then we describe the properties of Jeffreys divergence which makes it suitable for our problem.
For the rest of this section, we assume that the observation $x$ is fixed and $\pi(z) = p(z|x)$ is the target density function which we can evaluate up to a normalizing constant $\tilde \pi(z) = p(z, x)$ where $\pi(z) = \tilde \pi(z) / Z$ and $Z = p(x)$. We also define a proposal distribution with normalized density $\gamma_0(z)$ which is easy to sample and evaluate, such as the variational posterior $\gamma_0(z) = q_\phi(z|x)$.

\subsection{Recap of Annealed Importance Sampling}\label{sec:AIS}
To sample from the target distribution, AIS algorithm requires a sequence of distributions which change from the proposal distribution to the target. To match with our parameterization of the bridging distributions, we consider the generalization of this sequence to a continuously indexed path of density functions. 
With some abuse of notation, let $\gamma:[0,1]\times \R^d\to\R$ with mapping $(t,z)\mapsto \gamma(t,z)$ denote a path of density functions indexed by $t\in[0,1]$, with fixed endpoints $\gamma(0, \cdot) = q_\phi(\cdot|x)$ and $\gamma(1,\cdot)=\tilde \pi(\cdot)$. Assume the density functions have non-increasing support (i.e. $\supp(\gamma(t',\cdot)) \subseteq \supp(\gamma(t,\cdot))$ for $0\leq t\leq t'\leq 1$). Let $t_k=k/M$ for $k\in\{0, ..., M\}$ be the linear schedule in $[0,1]$ with which the path of density functions is discretized and let $\gamma_k(\cdot)=\gamma(t_k,\cdot)$. We assume that the $\gamma(\cdot, z)$ curves are flexible enough to absorb any  increasing mapping $\beta:[0,1]\to[0,1]$ (representing a discretization of $[0,1]$ at $\left(\beta(t_k)\right)_{k\in\{0,...,M\}}$). 

AIS algorithm starts by sampling the initial particles $z_0\sim \gamma_0(z)$ and evaluating the initial weights $w_0=1$. Then, particles are gradually moved in iteration $k$ using Markov kernel $\mk{k}(z_k|z_{k-1})$ such that the distribution of particles $z_M$ approaches the target distribution $\pi(z)$. The particle trajectory distribution in AIS has the joint density $\qfw = \gamma_0(z_0)\prod_k \mk{k}(z_k|z_{k-1})$. In each iteration a second Markov kernel $\mkrev{k}(z_{k-1}|z_k)$ is used to approximate the probability density of backward transition $z_k\to z_{k-1}$ such that the ratio $w_k(z_{k-1}, z_k) = \gamma_k(z_k)\mkrev{k}(z_{k-1}|z_k)/\gamma_{k-1}(z_{k-1})\mk{k}(z_k|z_{k-1})$ is well-defined on $\R^d$ and the importance weights are adjusted by multiplying $w_k(z_{k-1}, z_k)$ with the current weights. The final importance weights can be evaluated as
\begin{align} 
    w(z_{0:M}) = \frac{\tilde\pi(z_M)\qbw}{\qfw} = \prod_{k=1}^M w_k(z_{k-1}, z_k)\nonumber
\end{align}
where $\qbw = \prod_k \mkrev{k}(z_{k-1}|z_k)$. 

The optimal backward process $\qbw$ with zero variance estimator $\hat p^N(x)$ is given by inverse kernels
\begin{align}
    \overleftarrow{\mathcal T}^*_{k}(z_{k-1} | z_k) = \frac{\overrightarrow{q}_{\phi, k-1}(z_{k-1})\mk{k}(z_k|z_{k-1})}{\overrightarrow{q}_{\phi, k}(z_{k})} \nonumber
\end{align}
where $\overrightarrow{q}_{\phi, k}(z_{k})$ is the marginalization of $\overrightarrow{q}_{\phi, M}(z_{0:M})$ over all $z_m$ except $m\neq k$. However it is intractable to use $\overleftarrow{\mathcal T}^*_{k}(z_{k-1} | z_k)$ for sampling \cite{del2006sequential}. 

Alternatively, we can use a heuristically fixed $\left(\gamma_k\right)_k$ to guide the samples towards the target's basins of energy. One may choose $\mk{k}$ from popular MCMC kernels like Random Walk Metropolis-Hastings (RWMH), Hamiltonian Monte Carlo (HMC), Metropolis Adjusted Langevin Algorithm (MALA) or Unadjusted Langevin Algorithm (ULA) with $\pi_k$ as their stationary distribution  \cite{thin2020metflow,thin2021monte}. MH-corrected kernels typically do not admit a transition probability density. However, due to detailed balance $\gamma_k(z_k)\mk{k}(z_{k-1}|z_k) = \gamma_{k}(z_{k-1})\mk{k}(z_k|z_{k-1})$  the kernel $\mk{k}$ is reversible with respect to $\gamma_k$ and with backward kernel set to its reversal $\gamma_k(z_{k-1})\mk{k}(z_k|z_{k-1}) / \gamma_{k}(z_k)$ we can obtain well-defined importance weight updates $w_k(z_{k-1}, z_k) = \gamma_k(z_{k-1})/\gamma_{k-1}(z_{k-1})$ \cite{neal2001annealed}. When $\mk{k}$ is not $\gamma_k$-invariant, an approximate reversal kernel with same assumption may be constructed from $\mkrev{k}(z_{k-1}|z_k) = \mk{k}(z_{k-1}|z_k)$ while the weight updates preserves their original form \cite{thin2021monte}.

In another approach, reparameterizable forward and backward transition kernels can be used where the parameters are optimized to push samples from one predetermined bridging distribution to the next, which limits the complexity of the transition probability density considerably \cite{naesseth2018variational,wu2020stochastic,arbel2021annealed}. 

\section{Optimized Path Annealed Importance Sampling}\label{sec:motiv}
Typical implementations of AIS use predefined heuristics for the sequence $\left(\gamma_k\right)_k$ and the schedule $\beta$. Popular choices include the geometric average density path $\gamma(t, z) = \tilde\pi(z)^{\beta(t)}q_\phi(z|x)^{1 - \beta(t)}$ with linear $\beta(t) = t$ or geometric $\beta(t) = \alpha^{\log t}$ schedule, for some real $\alpha > 1$. Instead of focusing on optimization of the transition kernels for a fix annealing sequence, we propose to optimize the bridging distribution path $\gamma(t, \cdot)$ in a class of positive parametric functions. If the class is sufficiently large the schedule $\beta$ will be implicit in $\gamma$ and therefore we can fix the schedule to be linear and only focus on parameterizing the $\gamma(t, \cdot)$. We use Deep Neural Networks (DNN) to define our parametric density function path. We name our procedure Optimized Path Annealed Importance Sampling (OP-AIS).

\subsection{Parameterized AIS}\label{sec:param}
We opt to parameterize the time dependent path $\gamma$ as opposed to parameterizing each intermediary distribution individually. The main reason for this choice is that it results in a density path that is continuous in $t$ and allows parameters to be shared by intermediate distributions. We use a continuous mapping $u_\phi:\X\times\Z\times[0,1]\to \R$ with additional terms that adjust the boundary constraints at $t\in\{0,1\}$. We consider the following boundary adjustment which coincides with geometric average of the proposal and target distributions when $u_\phi\equiv 0$.
\begin{align}
        \log \gamma(t, z) = &u_\phi(x, z, t) \nonumber\\
            &+ (1-t)\left[-u_\phi(x, z, 0) + \log \gamma_0(z)\right] \nonumber\\
                &+ t\left[-u_\phi(x, z, 1) + \log \tilde \pi(z)\right]
\end{align}

If $u_\phi(x, z, t) \in(-\infty, \infty)$ is well-defined for all $x\in\X$, $z\in\R^d$ and $t\in(0,1)$ and the proposal is chosen such that $\supp(\gamma(1,\cdot))\subseteq \supp(\gamma(0,\cdot))$ then $\supp(\gamma(1,\cdot)) = \supp(\gamma(t,\cdot)) $ for all $t\in[0,1]$ and the importance weights updates are always defined using the common gaussian or invariant transition kernels. Notably, we can apply this parameterization to any arbitrary path of density functions by twisting the path with multiplication of an arbitrary positive function $\exp(-u_\phi(x, z, t))$ with correct boundary behavior.

\subsection{Optimization for the parametric distribution}
Given an unnormalized target distribution, Effective Sample Size (ESS) and its conditional extension are the de facto measure used to adaptively tune AIS \cite{del2006sequential,jasra2011inference,johansen2015towards,neal2001annealed}. ESS of $N$ parallel runs of AIS is given by
\begin{align}
    \text{ESS}(w^1, ..., w^N) =  \frac{\left(\sum w^j\right)^2}{N\sum (w^j)^2}\nonumber
\end{align}
and is expected to reflect the estimator variance odds with exact samples from $\pi(z)$ and importance sampling (\ref{eq:montecarlo}). ESS is sensitive to weight degeneracy problem that is prominent in AIS and other sequential importance sampling methods (when only a few particles contribute to the Monte Carlo approximate). However, high ESS does not determine if the particles have sufficient dispersion according to the target distribution. Maximization of this objective icorresponds to minimization of a consistent estimate of $\chi^2$-divergence between the sampling and extended target distribution, however typically results in insufficient dispersion of particles, therefore we omitted the results. %In fact, we can only trust high ESS values when sufficiently long annealing sequences are used to ensure that all the target modes are covered by a few samples. and 
See \cite{elvira2018rethinking} for further discussion on ESS and its defects. 

Grosse et. al. in \cite{grosse2015sandwiching} derived a bound on $\log Z$ by running the AIS algorithm in forward and backward directions. Original direction of AIS evaluates $\log \overrightarrow w = \log w(z_{0:M})$ which in expectation lower bounds $\log Z$.% i.e. from Jensen and Markov inequalities for any $c>0$ 
\begin{align}\label{eq:bdmclower}
    \E_{\overrightarrow{q}_{\phi, M}}[\log \overrightarrow w] &\leq \log Z%\nonumber\\
    %\text{Pr}(\log \overrightarrow w > \log Z &+ c) < e^{-c}
\end{align}
Drawing sample trajectories $z_{0:M}'$ from the reverse direction of AIS algorithm defined by the ancestral sampling in joint distribution $\pi(z_M')\overleftarrow{q}_{\phi, M}(z_{0:M-1}'|z_M')$ we can compute random variable $\log \overleftarrow w = \log w(z_{0:M}')$ which upper bounds $\log Z$ in expectation. % are available, random variable $\log \overleftarrow w = \log w(z_{0:M}')$ is a stochastic upper bound of $\log Z$ for any $c>0$.
%\begin{align}\label{eq:bdmcupper}
%    \E_{\pi\overleftarrow{q}_{\phi, M}}[\log \overleftarrow w] &\geq \log Z\nonumber\\
%    \text{Pr}(\log \overleftarrow w < \log Z &- c) < e^{-c}
%\end{align}
The gap between the two expectations in forward and backward processes is called the BiDirectional Monte Carlo (BDMC) gap and is equivalent to %the symmetric KL divergence between joint distributions of these processes, also known as 
Jefferys divergence between these distributions.
\begin{align}\label{eq:jefferys}
    \mathcal L^\text{BDMC}(x) &= \E_{\pi\overleftarrow{q}_{\phi, M}}[\log \overleftarrow w] - \E_{\overrightarrow{q}_{\phi, M}}[\log \overrightarrow w]\\
    &= \dkl(\pi\overleftarrow{q}_{\phi, M}||\overrightarrow{q}_{\phi, M}) + \dkl(\overrightarrow{q}_{\phi, M}|| \pi\overleftarrow{q}_{\phi, M}) \nonumber
\end{align}
%where $\dkl(f||g) = \int f(z)\log(f(z)/g(z))dz$ is the KL divergence between two distributions with density function $f$ and $g$.

This bound is also frequently used to assess the accuracy of the AIS estimator \cite{grosse2016measuring,wu2016quantitative,huang2020evaluating}. Unbiased estimation of this bound requires exact samples from the target distribution. Therefore, an approximation is made replacing the data samples with the synthesized samples by the generative model, where we have access to an underlying latent variable. This approximation results in more accurate log-likelihood estimation of samples closer to the model which affects the objectivity of the test. Assuming a long chain and close to perfect transitions the final particles from AIS forward process can be used to approximate this bound instead. However, with low computation budget this marginal distribution may be far off from the target especially in the initial optimization iterations. Although, the BDMC gap achieves a natural trade-off between minimization of the bias in ELBO bound and $\log$ empirical variance of the importance weights (Figure~\ref{fig:bench}), the two terms in $\mathcal L^\text{BDMC}(x)$ have contradicting gradients resulting in unstable optimization of hyperparameters with gradient based methods.

In order to achieve low variance importance weights we need to match the distributions of the forward and backward processes. It is common to minimize the inverse KL-divergence between these distributions similar to maximization of (\ref{eq:ELBO}) as this results in reducing the bias in (\ref{eq:bdmclower}). 
\begin{align}
    \Lais{} = \dkl(\overrightarrow{q}_{\phi, M}||\tilde\pi\overleftarrow{q}_{\phi, M}) = -\E_{\overrightarrow{q}_{\phi, M}}[\log \overrightarrow w] \nonumber
\end{align}
We can evaluate an unbiased estimate of inverse KL divergence using samples from $\overrightarrow{q}_{\phi, M}$. 

As was shown by authors of \cite{domke2018importance,maddison2017filtering} under some assumptions, the variance of importance weights can be controlled by the inverse KL-divergence between two distributions. In particular, with some adjustment to the AIS setting we have
\begin{align}
    \dkl(\overrightarrow{q}_{\phi, M}||\tilde\pi\overleftarrow{q}_{\phi, M}) \approx \frac{1}{2Z^2} \text{Var}_{\overrightarrow{q}_{\phi, M}}[w(z_{0:M})]\nonumber
\end{align}
Therefore, by minimizing the inverse KL-divergence we minimize the log marginal likelihood bias and the variance of the importance weights, simultaneously.
The equality holds when $\pi(z_k)\qbw = \qfw$ which results in zero importance weight variance and $\Lais{} = 0$. 

\section{Stochastic Optimization for OP-AIS}\label{sec:optim}
In order to evaluate the gradients of $\Lais{}$ with respect to the parameters $\phi$ we rely on the reparameterization of commonly used reversible transition kernels introduced by \cite{thin2020metflow} and their gradient estimation. The MH-corrected Markov kernel $\mk{k}(z_k|z_{k-1})$ is given by
\begin{align}
    \mk{k}(z_k&|z_{k-1}) = r(z_k|z_{k-1})\alpha_k(z_{k-1}, z_k)\nonumber\\
    &+ \int \left(1 - \alpha_k(z|z_{k-1})\right)r(z|z_{k-1})dz \delta_{z_{k-1}}(z_k)\nonumber
\end{align}
with $r(.|z_{k-1})$ denoting the conditional probability density of proposed particle in the $k$th iteration, $\alpha_k(z_{k-1}, z)=\left(1\land \frac{\gamma_k(z)r(z_{k-1}|z)}{\gamma_k(z_{k-1})r(z|z_{k-1})}\right)$ the acceptance rate at this iteration and $\delta_z$ denoting the Dirac measure at $z$. We assume that the proposal $r(z|z_{k-1})$ is reparameterizable with a auxiliary random variable $\eps_k\sim \eta(\eps)$ and transformation $z = T_k(\eps_k, z_{k-1})$ which is surjective in second argument. After transformation the proposal state is accepted with probability $\alpha_k(z_{k-1}, T_k(\eps_k, z_{k-1}))$ (denoted with $\alpha_k$ as a shorthand) and we set $z_k = T_k(\eps_k, z_{k-1})$. Otherwise, we keep the old state and set $z_k=z_{k-1}$. We determine the proper action by sampling a binary random variable $a_k\sim\text{Bern}(\alpha_k(z_{k-1}, T_k(\eps_k, z_{k-1})))$.

For instance, for a Random Walk kernel $\mk{k}$, we assume that $\eta(\eps)$ is the probability density function of standard normal distribution and $T_k(\eps, z) = z + \Sigma^{1/2}\eps$ for a positive definite covariance matrix $\Sigma\in\R^{d\times d}$. We refer the readers to the derivation of transformations corresponding to HMC, MALA and ULA by \cite{thin2020metflow} for further details.

Using the above reparameterization we get
\begin{align}
    \Lais{} = -\E_{\gamma_0(z_0)\prod_k \eta(\eps_k)\prod_k\zeta_k(a_k|z_0, \eps_{1:k}, a_{1:k-1})}\left[\log w\right]\nonumber
\end{align}
for $\zeta_k(a^k|z_0, \eps_{1:k}, a_{1:k-1}) = \alpha_k^{a_k}\left(1 - \alpha_k\right)^{a_k}$.

The gradient of $\Lais{}$ is given by
\begin{align}
    \nabla_\phi\Lais{} = 
    %&\E\left[\nabla_\phi\left(\left(\frac{w}{Z} - 1\right)\log w\right)\right]\nonumber\\
    & - \E\left[\nabla_\phi\log w\right]\nonumber\\
    %& + \sum_k\E\left[\left(\frac{w}{Z} - 1\right)\log w\nabla_\phi\log\zeta_k(a_k|z_0, \eps_{1:k}, a_{1:k-1})\right]\nonumber
    & - \sum_k\E\left[\log w\nabla_\phi\log\zeta_k(a_k|z_0, \eps_{1:k}, a_{1:k-1})\right]\nonumber
\end{align}

We can derive $\Lais{}$ approximation using the particle and importance weights generated from the AIS.
\begin{align}\label{eq:mclais}
    \nabla_\phi\Laishat{} = &- \frac{1}{N}\sum_{i=1}^N\nabla_\phi\log w^i\nonumber\\
    &- \underbrace{\frac{1}{N}\sum_{i=1}^N\sum_k\log w^i\nabla_\phi\log \zeta(a_k^i|z_0^i, \eps_{1:k}^i, a_{1:k-1}^i)}_{\nabla_\phi\hat{\mathcal L}^\text{score}(x)}\nonumber
\end{align}
%with $\omega^i=w^i/\sum_N w^l$ denoting the self normalized weights. 
$\nabla_\phi\Laishat{}$ is a strongly consistent estimator of $\nabla_\phi\Lais{}$ \cite{laubenfels2005feynman}.

Score function estimator is notorious for its high variance. It is a standard practice to omit $\nabla_\phi \mathcal L^\text{score}$ in parameter updates for more stable optimization. However, this results in biased gradient estimate which adversely effects our estimation in latent variable model experiements. %Therefore, in our experiments we don't consider this term as well.
To reduce the high variance of $\nabla_\phi\hat{\mathcal L}^\text{score}(x)$ we can employ the variation reduction techniques \cite{naesseth2017reparameterization,ranganath2014black}. For instance, in our experiments we use the common particles and weights to estimate the two expectations in $\Lais{}$ and the one-out control variates proposed by \cite{mnih2016variational} to substitute $\nabla_\phi\hat{\mathcal L}^\text{score}(x)$ with
\begin{align}
    \sum_{i=1}^N \left[\frac{NR^i - \sum_jR^j}{N-1}\right]\sum_k\nabla_\phi\log \zeta_k(a_k^i|z_0^i, \eps_{1:k}^i, a_{1:k-1}^i)\nonumber
\end{align}
where $R^i = -\log w^i/N$.

\begin{algorithm}[!htb]
    \SetKwInOut{Input}{input}
    \SetKwInOut{Output}{output}
    
    \Input{Target $\tilde\pi$, proposal density $\gamma_0$, and $\eta$}
    \Output{Parameters $\phi$}
    \For{$K$ epochs}{
        Set $\Delta \gets 0$\\
        \For{$i = 1$ to $N$}{
            Set $\text{logw}^i \gets 0$, $\text{log}\zeta^i \gets 0$\\
            Draw $z^i\sim q_0(z)$ \\
            Draw $\eps_{1:M}^i\sim \prod_{k} \eta(\eps_k)$ \\
        }
        \For {$k = 1$ to $M$}{
            \For {$i = 1$ to $N$}{
                Set $\text{logw}^i \gets \text{logw}^i + \log \frac{\gamma_k(z^i)}{\gamma_{k-1}(z^i)}$\\
                Draw $a_k^{(n)}\sim \text{Bern}(\alpha_k)$\\
                Set $\text{log}\zeta^i \gets \text{log}\zeta^i + \log \zeta(a_k^i|z_0^i, \eps_{1:k}^i, a_{1:k-1}^i)$\\
                Set $z^i \gets T_k(\eps_k^i, z^i)^{a_k^i}(z^i)^{1-a_k^i}$\\
            }
        }
        Get $R^i$ for $i\in[N]$\\
        $\phi \gets \phi - \kappa \nabla_\phi \Laishat{}$
    }
\caption{Optimization of parameterized AIS}\label{algo}
\end{algorithm}
Pseudocode of the optimization algorithm is given in Algorithm (\ref{algo}) for completeness. 

\section{Literature Review}
We can interpret AIS as the discretization of the path sampling algorithm where the thermodynamics integral identity holds
\begin{align}
    \log Z &= \int_0^1 \E_{\gamma(t,z)/Z_t} \left[\frac{d}{dt}\log \gamma(t, z)\right]dt\nonumber
\end{align}
for $Z_t = \int \gamma(t,z) dz$. Gelman et. al. in  \cite{gelman1998simulating} derived the optimal path and scaling of the bridging distributions for minimum variance estimation. Evaluation of optimal path is intractable in general. %In \cite{calderhead2009estimating}, the authors provide a tuning mechanism for the annealing schedule in path sampling for the path between two Gaussian distributions and \cite{behrens2012tuning} extend it to more general family of target distributions.
In turn, authors of \cite{calderhead2009estimating,behrens2012tuning} provided closed form optimal path for special cases. 
Viewing the importance weight updates as the finite difference approximations, the authors in \cite{grosse2013annealing} derive the asymptotic bias of $\log w$ estimator and propose moment average path between members of exponential family. A variational representation of the annealing distributions minimizing weighted $\alpha$-divergences between proposal and target distributions are derived in \cite{grosse2013annealing,masrani2021q}. In \cite{kiwaki2015variational}, the authors minimize the asymptotic variance of estimator $\log w$ for a given parametric path of distributions.

The analysis in these works are derived based on the assumption of reaching equilibrium in each iteration and independence of particles. This is not the case when transition kernels are not invertable. On the other hand, exhaustive heuristic optimization to tune AIS hyperparameters is laborious \cite{salakhutdinov2008quantitative}. Other methods to adapt AIS hyperparameters include monitoring the acceptance rate in Markov transitions to adjust its parameters \cite{jasra2011inference,thin2021monte} and designing problem specific or heuristic intermediary distribution sequences \cite{jasra2011inference,salakhutdinov2008quantitative,masrani2021q,grosse2013annealing}.
On the contrary, while our method requires training before evaluation, we can amortize the overhead by parameter sharing and reduce the computation cost by using only a small subset of validation samples. As a consequence, we find an efficient and shorter annealing process for sampling. Our aim is to enable AIS to achieve high accuracy despite limited computation resources. 

Recently, Variational Inference and Monte Carlo methods have been combined to increase the flexibility of parametric variational family $\Q$ and benefit from the convergence properties of sampling methods. %A tighter lower bound %by averaging importance weights of multiple latent variable draws in Importance Weighted AutoEncoders (IWAE) 
%was proposed in \cite{burda2015importance}. 
Authors of \cite{salimans2015Markov,wolf2016variational,caterini2018Hamiltonian,hoffman2017learning,thin2020metflow} combine MCMC and VI by extending the space of latent variables $\left(z_k\right)_{k\in[M]}$. %These methods optimize ELBO and contrastive divergence based on distributions of the forward Markov process $q_\phi(z_0|x)\prod \T(z_i|z_{i-1})$ and an approximate backward process starting from the true posterior. Annealing the target distribution in each iteration allows the model to learn simpler transition in comparison to MCMC based sampling methods, realizing forward and backward process distributions which are closer to each other.% \cite{Salimans2015,Wolf2016,Thin2020}. 
A framework with parametric transitions for SMC algorithms and a surrogate lower bound ELBO are provided in \cite{gu2015neural,tom2017auto,naesseth2018variational}. For example, in \cite{maddison2017filtering}, the authors derive an approximation of the asymptotic variance of importance weights in Variational Filtering Objectives, and in \cite{wu2020stochastic,arbel2021annealed}, flows are used to overcome energy density barriers. A differentiable alternative to MH-correcting step in HMC transitions are proposed \cite{zhang2021differentiable}, while in \cite{geffner2021mcmc} the authors derive the importance weights resulting from unadjusted HMC process and its reversal. %Extending the latent space with uncorrelated momentum variables may raise the variance of estimation especially if momentum has a high refreshment rate. 

Our work in particular bares more similarity to MCVAE proposed by \cite{thin2021monte} among others as we use the same reparameterization and optimization discribed in Section \ref{sec:optim}. In MCVAE the authors use AIS with heuristic density path for low variance estimation of the log marginal likelihood during training of a latent variable model. Our goal here is to reduce the computation complexity of evaluation of an already trained model by optimizing the intermediary distributions in AIS algorithm and we argue that the learnt AIS parameters by MCVAE are not efficient since the model is constantly changing during training and the posterior approximator is known to overfit the model \cite{wu2016quantitative}.% by interleaving the MH corrected transitions with bijective deterministic Markov kernels and therefore, improve the independence of different states particles in a trajectory. In fact, improving variational encoder in VI with NF is equivalent to using an SNF with only a single bijective deterministic Markov kernel \cite{improvingVINF}.

\section{Experiments}\label{sec:exp}
\begin{figure}[!htb]
    \includegraphics[width=1\textwidth]{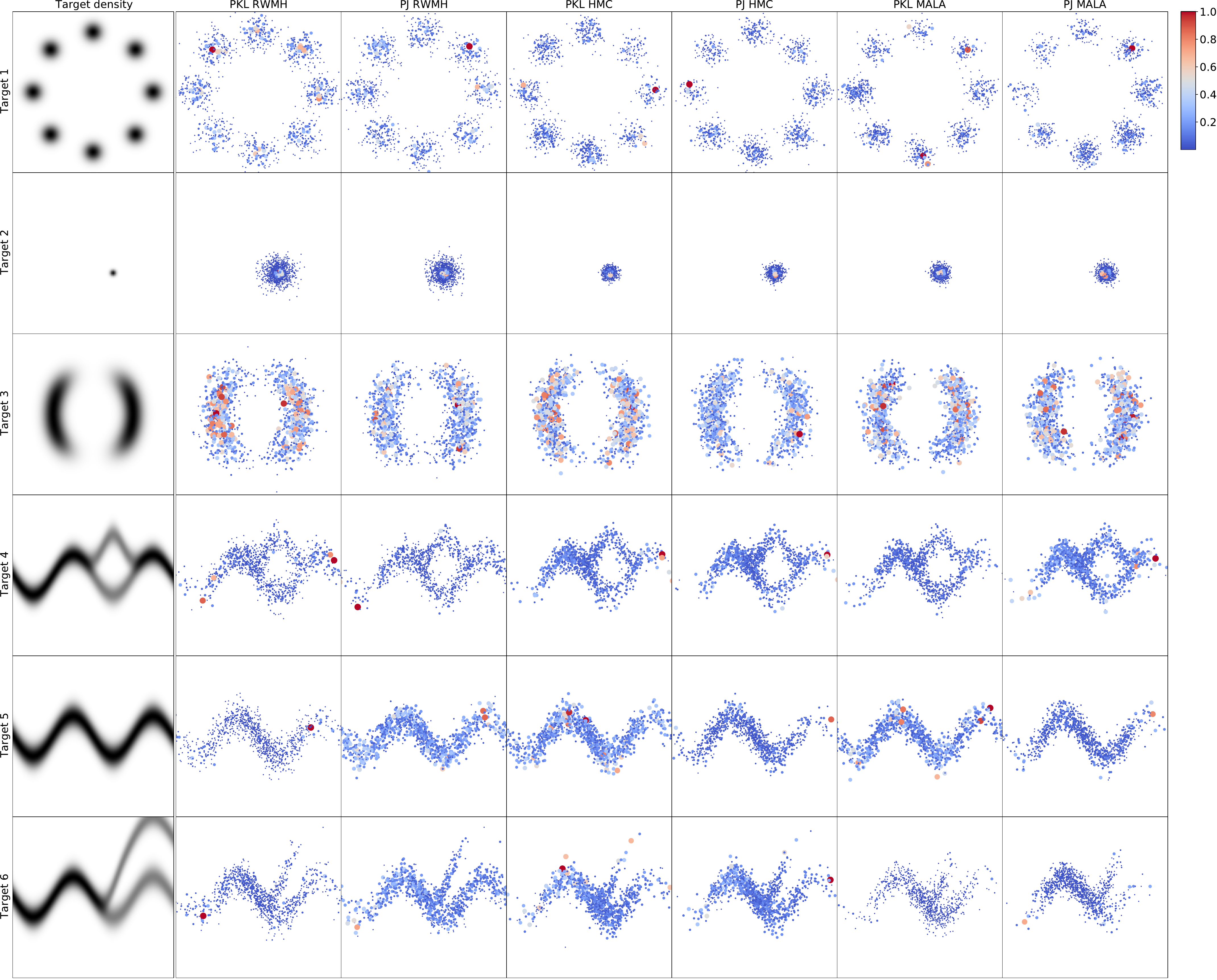}    
    \caption{Target distributions and particles of samplers with optimized path (trained with KL and Jefferys divergence) with RWMH, HMC and MALA transitions and $M=8$. Particles with larger weights are coded to show comparable scale to maximum weight in each plot. Larger dots with warmer colors show particles with larger weights, with red for maximum weight in each set and dark blue for particles with 0.1 of the maximum weight}
    \label{fig:samples_8}
\end{figure}

Here, we evaluate the proposed algorithm on complex synthetic 2d distributions which are often used to benchmark generative models \cite{rezende2015variational}. Target distributions are illustrated in Figure~\ref{fig:samples_8}. %We leave experiments on latent variable models and real datasets for future work. 
We compare our algorithm to vanilla AIS \cite{neal2001annealed} with geometric average path and geometric and linear schedule. We use RWMH, HMC and MALA transitions and normal proposal with learnable mean and diagonal covariance matrix. $M$ between 2-128 bridging distributions and $256$ samples are used for training.
$u_\phi$ is implemented with a DNN with one hidden layer with 4 dimensions and LeakyReLU(0.01) nonlinearty except at the output layer.

The parametric AIS is trained with Adam optimizer \cite{kingma2014adam} with learning rate of 0.03 and betas = (0.5, 0.999) for 100 epochs. The RWMH kernels have local normal proposal steps with learnable diagonal covariance matrix and the step size in all transition kernels are trained along with other parameters. We use cross validation to choose the step size in vanilla AIS from $\{0.01, 0.05, 0.1, 0.5, 1.\}$. The normalization constants are estimated using $N=4096$ samples. To act as the reference normalization constant we use vanilla AIS with $M=1024$, 10 MCMC steps per iteration with 3 different step sizes. All the code is written with Pytorch \cite{paszke2017automatic} and experiments are run on a GeForce GTX 1080 Ti. Code is available here: \url{https://github.com/shgoshtasb/op\_ais}.

In Figure~\ref{fig:samples_8} we illustrate particles generated from the samplers trained on each distribution with inverse KL divergence (PKL) and Jefferys divergence (PJ) color coded by their weight for $M=8$. % and $M=64$.
Warmer colors show particles with higher weights which are main contributors to the partition function estimation, while dark blue particles are less effective in the estimation. We observe that training with KL divergence results in smaller bias, while using Jefferys divergence objective results in more effective particles and lower variance estimation in Target 3-6.%Larger number of effective particles is also preserved with added bridging distributions, for $M=64$, while training with KL divergence results in worse coverage of the target in this case.
%Figure \ref{fig:annsamples} shows samples from the intermediate distributions  for parameterized AIS.% with their normalized importance weights encoded as their color. In the initial iterations the weights have high entropy and with higher values closer to the modes of the intermediate distributions. 
\begin{figure}[!htb]
    \includegraphics[width=1\textwidth]{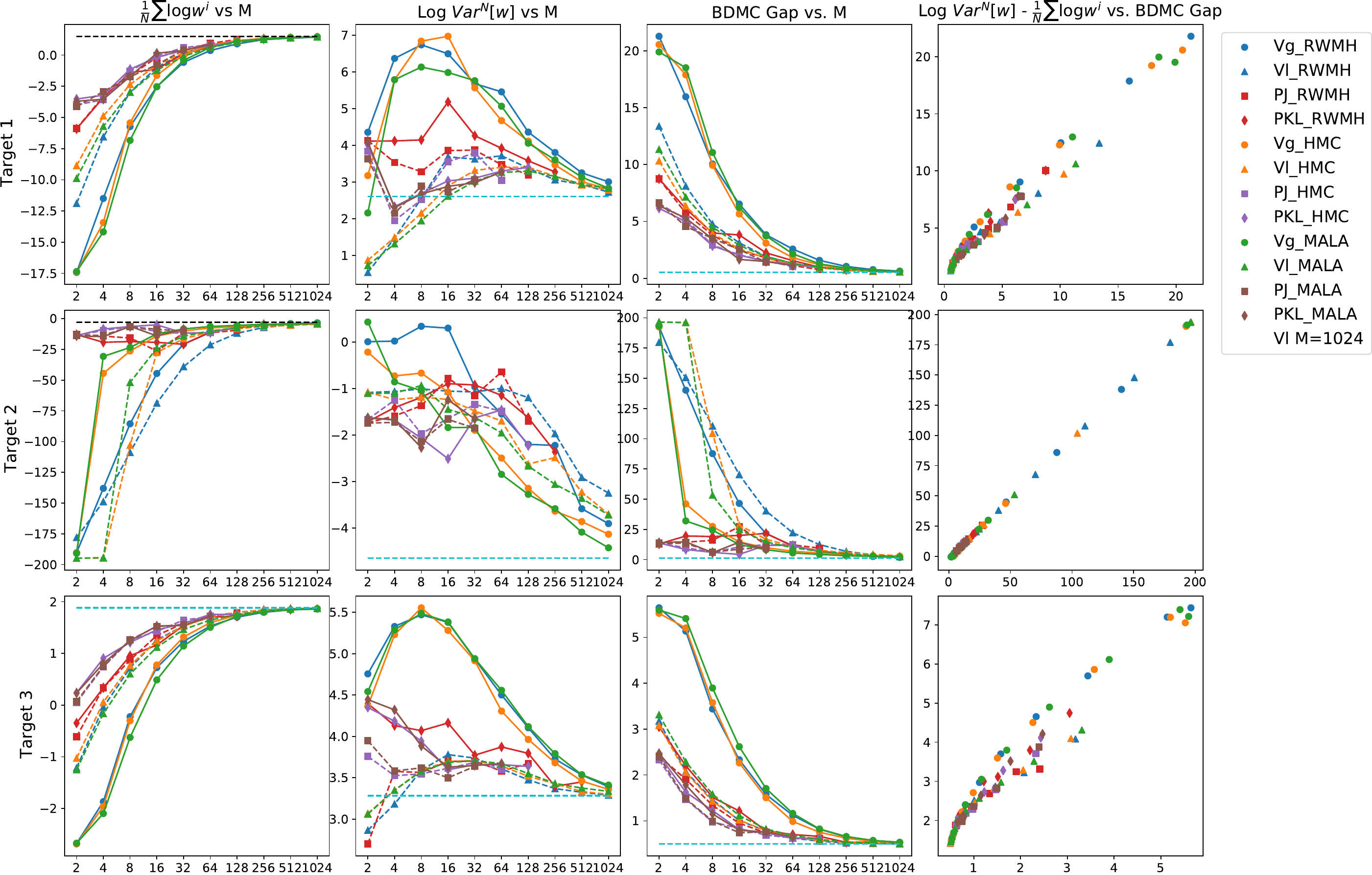}
    \caption{From left to right: $\log Z$ estimation with Eq.~\ref{eq:bdmclower}, log empirical variance of importance weights and the Jefferys divergence of forward and backward processes vs $M$%, $\log \text{VAR}^N[w] - \frac{1}{N}\sum\log w^j$ vs. BDMC Gap
    . (Vg, Vl for vanilla AIS with geometric and linear schedules, PJ and PKL for OP-AIS with Jefferys and KL-divergence optimization)}
    \label{fig:bench}
\end{figure}
Figure~\ref{fig:bench} shows the log partition function estimation and the empirical variance of importance weights for the mentioned samplers. We index the target distributions in the same order they appear in the Figure~\ref{fig:samples_8}. In most of the experiments, for small number of bridging distributions ($M\in\{2, 4, 8\}$) our method is able to improve over vanilla AIS algorithm with linear schedule and achieves slightly tighter lower bound in comparison to geometric schedule version. In comparison to training with inverse KL divergence, Jefferys divergence improves the variance of importance weights. we speculate this to be related to higher correlation of Jefferys divergence with weight variances.  

Interestingly, in multimodal and heavy tail distributions, empirical variance for small M remains below the variance of exhaustive AIS (Vl M=1024 in dashed Turquoise) and grows with $M$ %In this case samples from proposal need a lot of steps to get to the modes of target distributions and therefore, 
With only one MCMC step used in the experiments, particles ultimately cover a small area and don't reach the high density regions in the target distribution, resulting in low empirical weight variance despite the fact that the actual variance is much larger than our observation. This is specially problematic with the geometric schedule as the few bridging distributions are placed very close to the proposal and the particles are not encouraged to move far. In the low $M$ setting as it is unlikely to observe high weight particles, it is misleading to use ESS or even empirical $\chi^2$-divergence to tune the hyperparameters, whereas the inverse KL divergence provides better guidance for tuning. Increasing $M$ results in more dispersed particles and higher variance. For large enough $M$ as the particles can reach high density regions in the target, we observe that this variance decreases with $M$. Variance in parametric samplers is more steady and has a small growth in comparison due to the optimized choice of bridging distributions for each M.

\subsection{MNIST Dataset}
\begin{figure}[!htb]
    \begin{center}
    \includegraphics[width=1.\linewidth]{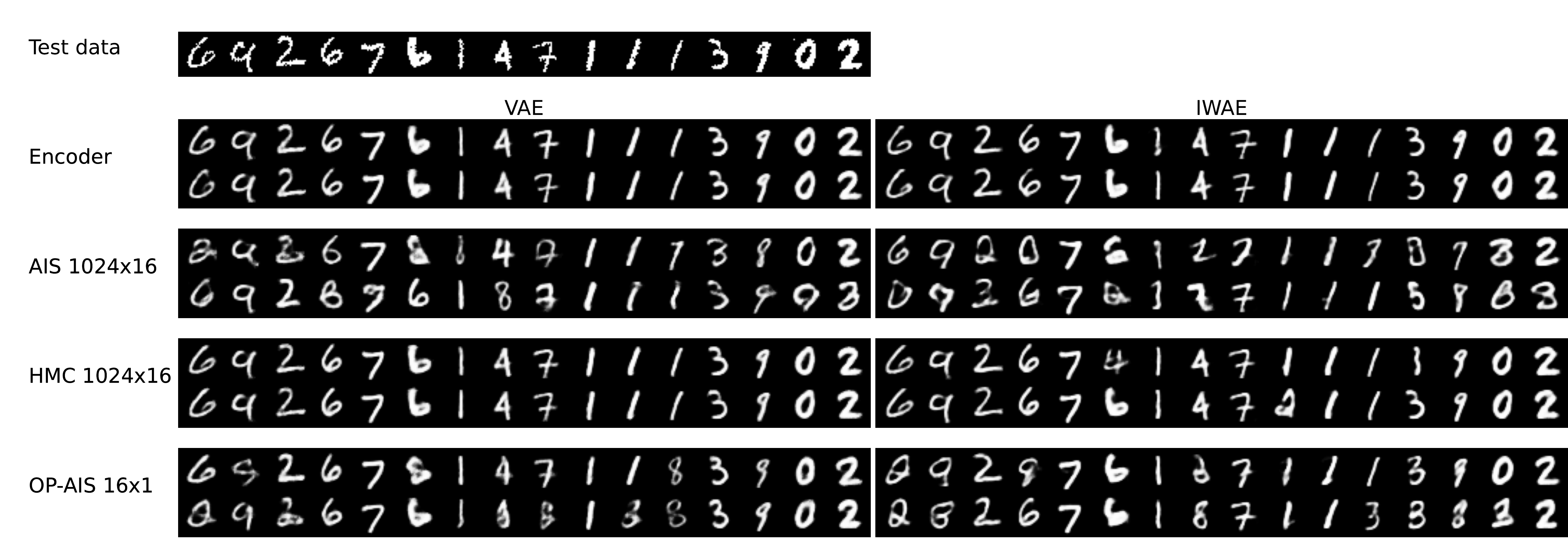}    
    \caption{VAE sample reconstruction with posterior samples}
    \label{fig:recon}
    \end{center}
\end{figure}
In this section, we assess the effectiveness of our method on a generative models based on similar DNN decoder architecture trained on MNIST dataset. We use the binarized MNIST dataset with Bernoulli observation model and data dequantization, rescaled to [0,1] \cite{salakhutdinov2008quantitative}. The decoder model we consider is the "Wide" architecture used in \cite{wu2016quantitative} with 50-1024-1024-1024-784 Fully-Connected (FC) layers, Tanh activations for hidden layers and Sigmoid for the output. This model is trained as a Variational AutoEncoder (VAE) \cite{kingma2013auto}, Importance Weighted AutoEncoder (IWAE) \cite{burda2015importance}, Generative Adversarial Network (GAN) \cite{goodfellow2014generative} and Bidirectional GAN (BiGAN) \cite{donahue2016adversarial}. We compare OP-AIS with $M\in\{8, 16\}$ and $N=16$ particles per data sample, HMC transition kernels with 1 leapfrog step to MCMC and Vanilla AIS with $M=1024$ and HMC transitions with 16 leapfrog steps. OP-AIS is trained with Adam optimizer with $0.01$ learning rate and betas=(0.5, 0.999) for 100 epochs on 1400 samples and validated on 600 samples for highest log likelihood estimation. Reported log-likelihood values are over remaining 8000 test samples. Here, we use a symmetric encoder with Tanh activation and Dropout(0.2) at each hidden layer to transform $x$, concatenat with $z$, and input to $u_\phi$ with 32-hidden units and LeakyReLU(0.01) activation. The encoder network is similar to the one used in \cite{wu2016quantitative} for the "Wide" decoder setup. The training and evaluation time of OP-AIS (2h38 for $M=16$ on average) is approximately similar to the sampling time required by the MCMC and Vanilla AIS algorithms (2h54 on average).
\begin{table}
    \caption{Negative Log likelihood estimation of generative models on binarized MNIST}\label{tab:likelihood}
    \centering
    \resizebox{\textwidth}{!}{
    \small{
    \sisetup{
        table-align-uncertainty=true,
        separate-uncertainty=true,
    }
    %% local redefinitions
    \renewrobustcmd{\bfseries}{\fontseries{b}\selectfont}
    \renewrobustcmd{\boldmath}{}

    \begin{tabular}{|c|c|cc|cc|}
    \hline
        \multicolumn{1}{|c|}{Model} & \multicolumn{1}{c|}{Encoder} & \multicolumn{2}{c|}{OP-AIS} & \multicolumn{2}{c|}{AIS} \\
        $M\times$steps & & $8\times 1$ & $16\times 1$ & $1024\times 16$ & $1024\times 10^{\text{Tuned}}$\\
        \hline

        VAE & $95.7$ & $216.7\pm29.8$ & \bfseries{193}$\pm$\bfseries{31.9} & $215.6\pm35.4$ & 87.8$\pm$5.8\\
        IWAE & $88.5$ & \bfseries{213.9}$\pm$\bfseries{27.3} & 221.2$\pm$34.1 & $235.4\pm35.8$ & 85.8$\pm$5.6\\
        GAN & & $805.9\pm94.7$ & \bfseries{785.4}$\pm$\bfseries{67.0} & $902.9\pm115.8$ & 415.3 $\pm$48.8\\
        BiGAN & & $900.5\pm124.6$ & $885.3\pm138.3$ & \bfseries{864.2}$\pm$\bfseries{98.1} & 373.2$\pm$44.6\\
    \hline
    \end{tabular}}}
\end{table}
                                                                                                                    
Table~\ref{tab:likelihood} shows the negative log marginal likelihood estimation using posterior samples from each sampling algorithm. The ground truth values are approximated using a more exhaustive version of AIS with 10 tuning iterations for each transition kernel on the 1400 separated data samples (AIS $1024\times 10^\text{Tuned}$) (8h12 average tuning + sampling time). Although the values of aquired by ground truth and encoder network (when available) are far from sampler estimations, estimates are more accurate than the sampling counterpart with similar computation time. We also observe that sample reconstruction with OP-AIS achieves similar samples to the test at least in style and are comparable to the other baselines in Figure~\ref{fig:recon}. 

\section{Conclusion}
AIS yields an unbiased estimation for any annealing path, schedule and transition operator. However, the variance of the Monte Carlo estimator changes considerably with different choices. We propose to optimize the intermediary distribution sequence in AIS algorithm by matching the distribution of the AIS particle trajectory to joint distribution of backward process with target distribution as its marginal. We compare commonly used measures for tuning and training AIS hyperparameters in their ability to reduce bias and variance of the estimation. Optimization of the annealing sequence improve the tightness of the lower bound and convergence rate on the 2D synthetic benchmarks and competes with more expensive heuristic sampling on latent variable models.

It is important to mention that log-likelihood has to be used with caution for model comparison, since a model with very good or bad generative sample quality or memorization may have high log likelihood \cite{theis2015note}. It is recommended to use other sample quality measures (e.g. \cite{yufrechet,salimans2016improved}) and model comparison methods (e.g. \cite{huang2020evaluating,sajjadi2018assessing}) along with AIS.

\subsubsection{Acknowledgements} This work is supported by funding from the European Union's Horizon 2020 research and innovation program under the Marie Sklodowska-Curie grant agreement No 813999 for this project.


\end{document}